\documentclass[runningheads]{llncs}

\usepackage{eccv}

\usepackage{eccvabbrv}

\usepackage{graphicx}
\usepackage{booktabs}

\usepackage[accsupp]{axessibility}  

\usepackage{multirow}

\usepackage{hyperref}

\usepackage{orcidlink}

\begin{document}

\title{Analysis of Hybrid Compositions in Animation Film with Weakly Supervised Learning} 

\titlerunning{Weakly Supervised Learning in Animation Film}


\author{Mónica Apellaniz Portos\inst{1}\orcidlink{0009-0007-2030-0720} \and
Roberto Labadie-Tamayo\inst{1}\orcidlink{0000-0003-4928-8706} \and
Claudius Stemmler\inst{2}\orcidlink{0009-0005-0101-4070} \and
Erwin Feyersinger\inst{2}\orcidlink{0000-0002-2366-5748} \and
Andreas Babic\inst{1}\orcidlink{0009-0008-8927-2593} \and
Franziska Bruckner\inst{1}\orcidlink{0000-0003-3455-8582} \and
Vrääth Öhner\inst{1}\orcidlink{0000-0002-5122-4165} \and
Matthias Zeppelzauer\inst{1}\orcidlink{0000-0003-0413-4746}}

\authorrunning{M.~Apellaniz et al.}

\institute{St. Pölten University of Applied Sciences, Austria \and
University of Tübingen, Germany
}
\maketitle

\begin{abstract}
  We present an approach for the analysis of \textit{hybrid} visual compositions in animation in the domain of ephemeral film. 
  We combine ideas from semi-supervised and weakly supervised learning to train a model that can segment hybrid compositions without requiring pre-labeled segmentation masks. We evaluate our approach on a set of ephemeral films from 13 film archives. Results demonstrate that the proposed learning strategy yields a performance close to a fully supervised baseline. On a qualitative level the performed analysis provides interesting insights on hybrid compositions in animation film. 
  
\end{abstract}

\section{Introduction}
\label{sec:introduction}



\begin{figure} [ht!]
    \centering    \includegraphics[width=\textwidth]{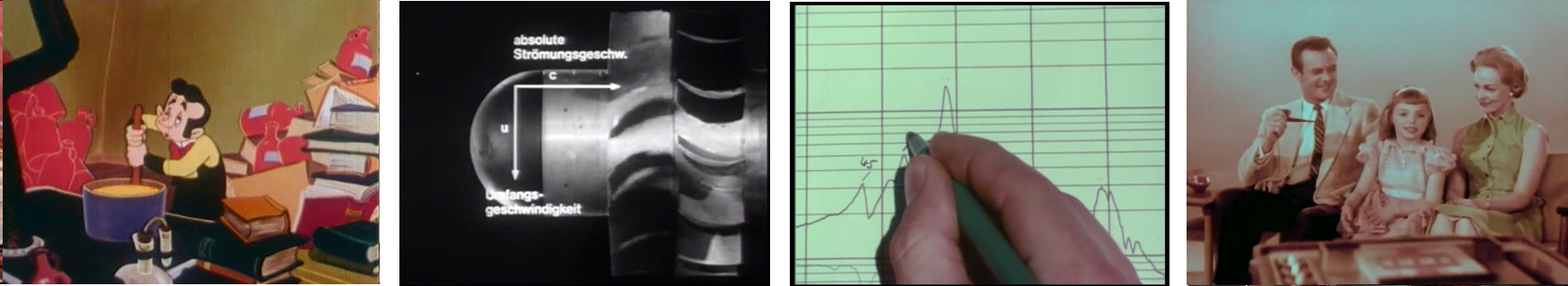}
    \caption{Four sample frames from our database (left to right): a frame from a fully animated sequence \cite{Fig1Animation}, two hybrid compositions combining photographic 
    and graphic content \cite{tibav:14363,tibav:10786}, 
    and a purely live-action frame \cite{Fig1Live}.}
    
    \label{fig:intoFig}
    \vspace{-15pt} 
\end{figure}

Animation is older than the medium film itself and has been incorporated into all kinds of formats since the beginning of film history. However, research on animation has mainly focused on its narrative, experimental and, more recently, documentary uses \cite{gros_animationstheorien_2016}. 
One context of animation that has been marginalized in research is its use in \textit{non-fiction films} with a recognizable utilitarian purpose, so-called \textit{ephemeral films}. Inside these films, there is a great variety of visual compositions. Some of the most 
intricate compositions are so-called ``\textit{hybrid}'' compositions, which combine live-action film and animated content, see \cref{fig:intoFig} for examples of hybrid and non-hybrid compositions, and the supplementary material for a more detailed introduction to 
hybrid compositions.

While film scholars have recently started to explore ephemeral films in general, animation in ephemeral film has rarely been studied. 
A quantitative analysis of animation styles, especially of hybrid compositions is currently missing. We have compiled a corpus of ephemeral films from 13 film archives located in Germany and Austria.
This dataset enables for the first time a quantitative and automated analysis of the visual style of animation in ephemeral films, particularly of hybrid compositions.

In this paper, we focus on the automated analysis of hybrid compositions in ephemeral film, because they are one of the most complex and artistically interesting types of visual composition for film scholars. 
Thereby, we focus on the \textit{frame level}, as a hybrid composition is independent of motion. 
The primary challenge for our approach is that there are no spatial annotations for hybrid content in our corpus, and its manual labeling is too time-consuming in practice. 
To tackle this challenge, we propose a segmentation approach for that requires only a minimum of manually provided labels at the \textit{sequence level}. We integrate and combine ideas from weakly supervised learning (using global labels to solve a task at a local level) and semi-supervised learning (using proxy/pseudo labels to facilitate the learning). 

More concretely, we first learn a proxy task, \ie, the classification of photographic vs. non-photographic content for which training labels can easily be generated at sequence level. We employ this classifier to generate  segmentation masks that guide the training of a segmentation model for hybrid material. This completely frees our approach from requiring labels for hybrid material. In particular, we investigate the following research questions (RQ):
    \begin{itemize}
        \item \textbf{RQ1}: How well can we differentiate photographic from non-photographic visual content in ephemeral film automatically on a frame level? 
        \item \textbf{RQ1.1}: Are there boundary cases that are undecidable? If yes, how do they look like and how is the decision grounded by the model?
       \item \textbf{RQ2}: Can we leverage the trained classifier from RQ1 as proxy to guide the training of a spatial segmentation method for hybrid compositions?
       \item \textbf{RQ2.1}: Can we leverage visual transformer architectures to obtain local predictions on frames and derive a spatial segmentation from these predictions?
       \item \textbf{RQ2.2} How effective is the inclusion of proxy labels in our semi-supervised methodology? How close can we come to a supervised baseline?  
    \end{itemize}

We evaluate our approach on a diverse film corpus with a wide range of animation techniques and styles, produced over a long period of time. Our results show that the generated proxy masks are of surprisingly good quality and serve the semi-supervised training of our segmentation method well. Our main contributions are (i) a first quantitative investigation of hybrid visual compositions in animation films; 
(ii) a combined weakly and semi-supervised approach for training a segmentation model; and (iii) a new strategy to obtain spatial segmentation masks from a visual transformer by learning class-wise centroid representations at patch-level. 

\section{Related Work}
\label{sec:rw}

\textbf{Animation in ephemeral film.} Ephemeral film is one of several overlapping terms used to describe a large group of films with slightly different scope and connotations. Alternatives include the German term ``Gebrauchsfilm'' \cite{Hediger05}, its English translation utility film \cite{Hediger09,Bonah18}, non-theatrical film \cite{Slide92}, and useful cinema \cite{Acland11}. 
Compared to fiction and documentary film, there is only a small but growing body of research on these films. Typically, research takes the form of qualitative analyses focusing on specific aspects of smaller corpora, as in \cite{Hediger09,Bonah18}. Apart from advertising \cite{Cook19,Forster13}, animation is, barring sporadic exceptions \cite{Ostherr18}, only cursorily treated. A study on animated and hybrid elements in 54 German short films from 1958 to 1969 
has been contributed in 
\cite{filzmaier_grafic_2017, eckel_oberhausen_2018}. 
One field 
that overlaps with ephemeral films is that of animated documentaries \cite{ehrlich_animated_2013, honess_roe_animated_2013}. Documentary contexts of animation range from animated infographics in journalism and explanatory videos to documentaries with animated sequences and completely animated documentary films 
\cite{bruckner_visuelle_2017, gros_animationstheorien_2016, honess_roe_animated_2013}. To summarize, animation in ephemeral film has been little studied, with a few exceptions that are qualitative in nature, and research has generally been limited in its quantitative scope.

\textbf{Automated analysis of film style and animation film.} Research on automated film analysis has evolved in different research initiatives~\cite{flueckiger2017digital,heftberger_digital_2018,rushmeier2015examples,zaharieva_film_2011}, but animation is still underrepresented. 
Pioneering projects on automated film analysis were \textit{Digital Formalism}, focused on historic documentaries and visual composition in monochromatic film~\cite{zaharieva_film_2011}, \textit{Cinemetrics}, focused on shot lengths and cut frequency~\cite{tsivian2009cinemetrics}, and \textit{Videana}, focused on automated shot segmentation and camera motion detection~\cite{ewerth_videana_2009}. More recent developments are \textit{VIAN}~\cite{fluckiger2020digitale}, with a focus on visual color composition, and \textit{VIVA}~\cite{muhling2022viva}, which enables similarity search and visual concept detection. 
While a rich body of research analyzes and extracts basic video attributes
~\cite{vijayakumar_study_2012}, there is less work on the analysis of \textit{stylistic} attributes of film. The statistical analysis of film style goes back to Salt~\cite{salt_statistical_1974}, who proposed to quantify film characteristics by statistical measures and whose approach has been adopted by many researchers
~\cite{alvarez_influence_2019,adams2001automated,adams2002formulating}. 
Further work on visual composition and film style include the retrieval of motion composition~\cite{zeppelzauer_retrieval_2011}, visual composition~\cite{mitrovic_retrieval_2011}, montage patterns~\cite{zaharieva_recurring_2012}, and audio-visual film montage~\cite{zeppelzauer_cross-modal_2011}. 

While most related research on automated film analysis focuses on feature film, documentaries, and news broadcasts, only little work is concerned directly with animation. Exceptions are Ionescu et al.~\cite{ionescu_improved_2006}, detecting cuts and transitions in animated content. Furthermore,~\cite{ionescu_color-action_2011} classifies animated and non-animated content based on 
stylistic features. Similarly,~\cite{glasberg_cartoon-recognition_nodate} differentiates cartoons from live-action via content-based analysis. 
A color-independent approach for the classification of animation is presented in~\cite{zumer_color-independent_2018} and an approach based on motion content in~\cite{roach_motion-based_2001}. The authors of~\cite{ionescu_fuzzy_2008} present an approach for the characterization of color  
in animated sequences to retrieve sequences of similar style and appearance. 
The authors of~\cite{pais_animated_2012} introduce a method for the classification of animation genre and, therefore, fuse textual and visual information.

Most related to our work are the aforementioned works on the differentiation of live-action and animated content, which employ pre-segmented video with sequence labels for supervised training, and mostly extract traditional content-based sequence features (color, motion, shape). 
In our own preliminary experiments, we observed that color is in general not a discriminatory feature for animation (at least not in ephemeral film). 
We refrain from hand-crafted features and build upon learned (hierarchical) representations from visual transformer (ViT) models~\cite{dosovitskiy2020image}. In contrast to the works mentioned above, we not only focus on differentiating live-action and animated sequences (on the global sequence level) but on the \textit{concurrent and intertwined composition of live-action and animated sequences within individual frames}. In absence of ground truth, we refrain from supervised learning and propose a weakly and semi-supervised~\cite{zhou2018brief} strategy to train a segmentation method. 

\section{Methodology}
\label{sec:method}

    In the following, we present our approach towards the segmentation of hybrid visual composition. First, we introduce some basic terminology 
    used in our work.

    \subsection{Terminology}
    A hybrid composition concurrently combines different content types in one frame, namely \textit{photographic} content (short: P) and \textit{non-photographic} content (short: NP). 
    With the term photographic, we refer to visual content that is initially created with photography, \ie, a technical reproduction of a real-world scene using photosensitive surfaces. By contrast, the term non-photographic refers to visual content that is initially created by other means, \eg, written text or drawings; thus, a photo (or film) of a drawing would still be considered a non-photographic element.
    Together, these two content types can be considered the visual building blocks from which any image is created. 
    Furthermore, we differentiate between \textit{homogeneous} and \textit{heterogeneous} frames. Homogeneous frames contain either purely photographic or purely non-photographic content. 
    Heterogeneous frames are hybrid visual compositions that contain both photographic and non-photographic content.

    \subsection{Method Overview}
\begin{figure}[h!]
        \centering
        \includegraphics[scale=0.26]{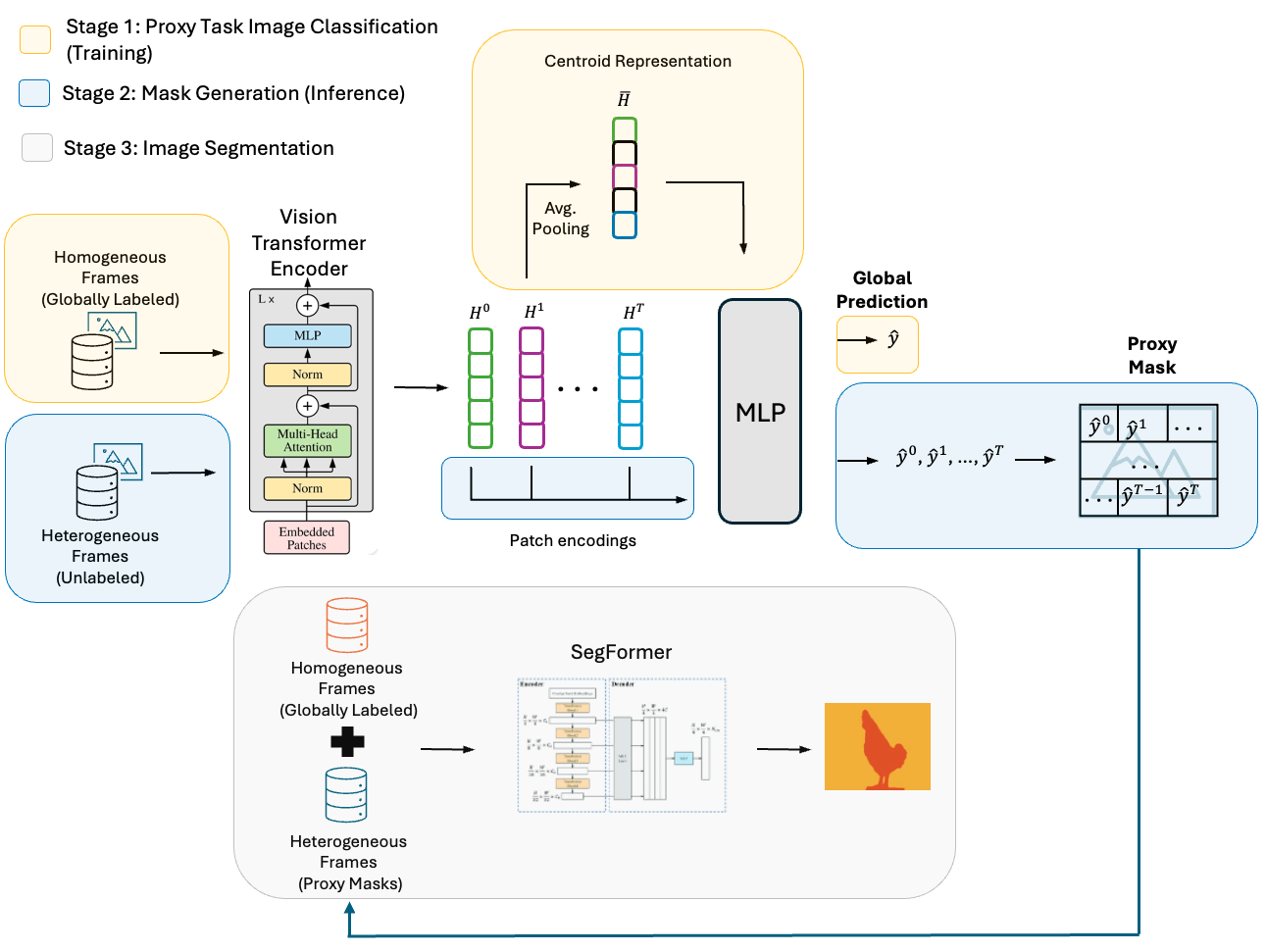}
        \caption{Proposed overall approach including proxy task learning (stage 1, yellow), segmentation mask generation (stage 2, blue) and image segmentation (stage 3, grey).}
        \label{fig:architecture}
        \vspace{-10pt}
    \end{figure}

    \figureautorefname~\ref{fig:architecture} provides an overview of our weakly and semi-supervised approach for the segmentation of hybrid visual compositions.
    Our approach consists of 3 stages: In stage 1, we train a proxy task that, in absence of segmentation ground truth for hybrid compositions, provides guidance in the later segmentation. The proxy task is a binary classification of homogeneous frames into the classes P and NP.   
    Stage 2 uses the proxy classifier to generate local segmentation masks for heterogeneous frames (called proxy masks). Proxy masks indicate regions corresponding to P and NP content in heterogeneous frames. They are obtained by weak supervision (from global labels). 
    In stage 3, both homogeneous frames (with homogeneous masks) and heterogeneous frames (with proxy masks) are used to train the segmentation model in a semi-supervised fashion. The segmentation model is finally evaluated on expert ground truth. 
    
    \subsection{Stage 1: Image Classification (Proxy Task)}\label{subsec:method_classification}

    For the modeling and classification of P vs. NP content, we  finetune a visual transformer (ViT) model~\cite{dosovitskiy2020image}, pretrained with the DINOv2 method~\cite{oquab2024dinov}\footnote{Pretrained weights available from \url{https://huggingface.co/timm/vit_base_patch14_dinov2.lvd142m}}. 
    
    Transformer models are able to learn complex features and share information globally, which has placed them as state-of-the-art architecture in many tasks in computer vision. Besides this, what makes pre-trained transformer-based models especially convenient for our approach, in particular considering stages 2 and 3, is the way in which the input signals are processed, namely as sequence of patches. This yields local information that is necessary to generate proxy segmentation masks in the following stages. 
    
    Given an image $I$ with height $H$, width $W$ and $C$ channels, $I \in \mathbb{R}^{H\times W\times C}$, it is convolved with learned filters $f^r \in \mathbb{R}^{p\times p}$ with a stride of $p$, $r = 1...d$. This produces a mapping where patches with window size $p$ ($p = 14$ in our case) are projected into a $d$-dimensional space. This embedding is flattened and processed by stacked transformer blocks via multi-headed self-attention layers. This equates to splitting image $I$ into non-overlapping patches of size $p\times p$, as:
    \vspace{-15pt}
    \begin{align*}
        I'_{t_{i,j}} = I[i*p : (i + 1)*p, j*p : (j + 1)*p]; ~~~~i \le \frac{H}{p}, j \le \frac{W}{p},
    \end{align*}
    where $I'_{t_{i,j}} \in \mathbb{R}^{ p \times p \times C}$ represents the patch at position $(i,j)$.   Next, $I'$ is unrolled into a sequence of 3-dimensional patches $X\in \mathbb{R}^{N \times (p \times p \times C)}$ with $N = \frac{WH}{p^2}$ and each patch is convolved as follows: 
    \begin{align*}
        Z^t_r = \sum_{(m,n)}^{(p,p)}\sum_k^C X^t_{(m,n,k)}*f^r_{(m,n,k)},
    \end{align*}   
    where $Z^t_r$ represents the $r^{th}$ component in the projection of the $t^{th}$ patch given by filter $f^r$.     For this sequence of projections $Z^t$, we obtain a corresponding sequence of hidden states $H^t$ as the output of the transformer backbone represented in \figureautorefname~\ref{fig:architecture} as ``patch encodings'' $H^0$, $H^1$ ,.., $H^T$, with $T$ the number of patches.
    
    In the case of the ViT model, additionally to the patches from $Z^t$, a class token [CLS] is fed into the stack of transformer blocks. The hidden state of this token is the one employed in the pretraining (and commonly in the fine-tuning) to assign a class to the input image. Since we aim to generate local image segmentations in the form of labels for each patch, we base the classification of an image on the average of the vectors in $H^t$ for each class (see ``Centroid Representation''  in \figureautorefname~\ref{fig:architecture}). The encoding $\Bar{H} = N^{-1}*\sum_t H^t$ provides us with the centroid of the region in the embedding space where patches of homogeneous images, from either classes P or NP, are located. By optimizing the network for $\Bar{H}$, we force the model to adapt to these regions. The centroid-based approach enables to optimize the model for patch-based predictions for the two target classes. The [CLS] token, in contrast, is optimized at the global image level, which we expect to be suboptimal when patch-level output of the model is required.
    

    On top of the transformer backbone, a Multilayer Perceptron (MLP) is stacked (see  \figureautorefname~\ref{fig:architecture}). This MLP is composed of a single ReLU-activated dense hidden layer, which is in charge of condensing the features from  $\Bar{H}$ into a 128-dimensional space before feeding them into a classification layer to predict $\hat{y}$.
    
    
    

    \subsection{Stage 2: Mask Generation}\label{subsec:method_dataS}

    Once the transformer-based classifier is fine-tuned, for each input image $I$, we can generate a coarse (\ie, patch-level) mask 
    by predicting one label for each of the projections in $H^t$ by the MLP directly without aggregating them into $\Bar{H}$. This results in one predicted label per patch for either class P or NP,  $\hat{y}^0, \hat{y}^1, ..., \hat{y}^T$. To obtain a spatial segmentation mask (proxy mask) for image $I$, each of the predicted labels is associated with the corresponding region covered by the respective patch in the input image; see blue boxes in \figureautorefname~\ref{fig:architecture}.



    
    \subsection{Stage 3: Image Segmentation}\label{subsec:method_segmentation}
    
    For segmentation we rely on a pre-trained SegFormer \cite{xie2021segformer} model. SegFormer is a segmentation model that leverages a transformer architecture in the encoder part to capture long-range dependencies within the input signal. This method provides high accuracy in the segmentation of complex regions that might have large and complex shapes, as can easily be the case in hybrid visual compositions. 
    
    The overall architecture of SegFormer consists of an encoder based on the Swin transformer \cite{liu2021swin} and a simplistic decoder based on an MLP. The Swin encoder enables the extraction of hierarchical and multi-scale mappings of the image to generate detailed segmentation maps. These multi-scale mappings are upscaled via bilinear interpolation in the decoder. Afterwards, the mappings are projected via a dense layer and concatenated to generate the output label for each pixel. 
    Each pixel will be classified into either class P or NP. 

    We finetune the SegFormer in a semi-supervised way. We use first homogeneous frames of class P and NP for which (homogeneous) segmentation masks can be generated automatically from the global image label. Second, we include heterogeneous frames (in the training and validation set) for which we pass the proxy masks to the SegFormer. The proxy masks represent coarse segmentation masks that guide the learning of the segmentation model.

\section{Data Corpus and Experimental Setup}
\label{sec:experimentalSetup}
\subsection{Data Corpus}

The digitized film data has been contributed by 13 film archives located in Germany and Austria.
Archives include publicly funded organizations 
as well as corporate archives. 
Most of the films were produced in these countries 
between 1945 and 1989. In total, there are more than 2000 films, ranging in length from short advertisements to long recordings of television programs. The visual quality of the digitized films varies greatly from archive to archive and film to film. The data includes advertisements, newsreels, educational, scientific, and corporate films in color and black and white. Animation techniques such as cutouts, stop-motion, and cel animation are well represented, while modified base, pixilation, and other animation techniques often associated with experimental filmmaking are rather scarce. Approximately one-third of the corpus has been annotated at \textit{sequence level}, where a sequence can consist of multiple shots. 
In this manner, pure live-action sequences, still images (static sequences), and animated sequences have been marked.
Finally, in case we observed hybrid sequences in the annotation process, they were annotated.\footnote{Due to legal restrictions, data can not be made publicly available.} 


\subsection{Dataset Compilation}
\label{subsec_DatasetPreparation}
In a first step, we compiled a dataset for the proxy classification task containing homogeneous frames, \textit{purely} photographic (P) and \textit{purely} non-photographic (NP) segments ($D1$), see \cref{tab:datasets} for an overview. The challenge thereby is to effectively identify such purely P or NP images from the corpus. From the available sequence-level annotations, we can almost automatically compile such a dataset. The homogeneous P class includes frames from segments originally annotated as ``live-action'', and ``photographic still image''. Instead, the homogeneous NP class is composed of frames from specific animation segments, such as ``drawn animation'', ``modified base painted on paper'', ``drawn cameraless animation'', and ``full and limited cel''; the NP class also includes ``graphic still image'' frames. 


Analogously, we compile a dataset $D2$ which serves our segmentation experiments. The first half of this dataset contains a randomly selected subset of homogeneous P and NP frames from $D1$. The rest of $D2$ contains heterogeneous frames that show hybrid visual compositions. This dataset includes segmentation masks in the form of  homogeneous masks for homogeneous frames ($D2_{NP}$ and $D2_{P}$), proxy masks generated via the proxy task for heterogeneous content ($D2_{Proxy}$), and precise manually labeled masks ($D2_{M}$) for evaluation.


\vspace{-10pt}
\begin{table}[h!]
\centering
\resizebox{\columnwidth}{!}{%
\begin{tabular}{|cc|ccccc|c|cccccccc|}
\hline
\multicolumn{2}{|c|}{\textbf{Task}} &
  \multicolumn{4}{c|}{\textbf{Classification}}  &
  \multicolumn{1}{c|}{\textbf{Total of}} & 
  &
  \multicolumn{7}{c|}{\textbf{Segmentation}} &
  \multicolumn{1}{c|}{\textbf{Total of}} \\ \cline{3-6} \cline{9-15}
\multicolumn{2}{|c|}{\textbf{Class}} &
  \multicolumn{2}{c|}{\textbf{Non-Photog.}} &
  \multicolumn{2}{c|}{\textbf{Photog.}} &
  \multicolumn{1}{c|}{\textbf{Samples}} & &
  \multicolumn{2}{c|}{\textbf{Non-Photog.}} &
  \multicolumn{3}{c|}{\textbf{Heterogeneous}} &
  \multicolumn{2}{c|}{\textbf{Photog.}} &
  \multicolumn{1}{c|}{\textbf{Samples}} \\ \cline{1-7} \cline{9-16}
\multicolumn{1}{|c|}{\multirow{3}{*}{\textbf{Subset}}} &
  \textbf{Train} &
  \multicolumn{1}{c|}{$D1_{NP}^{train}$} &
  \multicolumn{1}{c|}{1895} &
  \multicolumn{1}{c|}{$D1_{P}^{train}$} &
  \multicolumn{1}{c|}{2143} &
  \multicolumn{1}{c|}{4038} & &
  \multicolumn{1}{c|}{$D2_{NP}^{train}$} &
  \multicolumn{1}{c|}{158} &
    \multicolumn{1}{c|}{$D2_{Proxy}^{train}$} &
    \multicolumn{1}{c|}{$D2_{M}^{train}$} &
  \multicolumn{1}{c|}{158} &
  \multicolumn{1}{c|}{$D2_{P}^{train}$} &
  \multicolumn{1}{c|}{158} &
  \multicolumn{1}{c|}{474} \\ \cline{2-7} \cline{9-16}
\multicolumn{1}{|c|}{} &
  \textbf{Val} &
  \multicolumn{1}{c|}{$D1_{NP}^{val}$} &
  \multicolumn{1}{c|}{396} &
  \multicolumn{1}{c|}{$D1_{P}^{val}$} &
  \multicolumn{1}{c|}{473} &
  \multicolumn{1}{c|}{869} & &
  \multicolumn{1}{c|}{$D2_{NP}^{val}$} &
  \multicolumn{1}{c|}{110} &
    \multicolumn{1}{c|}{$D2_{Proxy}^{val}$} &
    \multicolumn{1}{c|}{$D2_{M}^{val}$} &
  \multicolumn{1}{c|}{99} &
  \multicolumn{1}{c|}{$D2_{P}^{val}$} &
  \multicolumn{1}{c|}{110} &
  \multicolumn{1}{c|}{319} \\ \cline{2-7} \cline{9-16}
\multicolumn{1}{|c|}{} &
  \textbf{Test} &
  \multicolumn{1}{c|}{$D1_{NP}^{test}$} &
  \multicolumn{1}{c|}{430} &
  \multicolumn{1}{c|}{$D1_{P}^{test}$} &
  \multicolumn{1}{c|}{438} &
  \multicolumn{1}{c|}{868} & &
  \multicolumn{1}{c|}{$D2_{NP}^{test}$} &
  \multicolumn{1}{c|}{94} &
    \multicolumn{2}{c|}{$D2_{M}^{test}$} &
  \multicolumn{1}{c|}{92} &
  \multicolumn{1}{c|}{$D2_{P}^{test}$} &
  \multicolumn{1}{c|}{94} &
  \multicolumn{1}{c|}{280} \\ \hline
\end{tabular}%
}
\caption{Overview of the two datasets: $D1$ for the proxy classification task, and $D2$ for the segmentation task. Subscript ``Proxy'' means that the dataset contains proxy masks for the images. Subscript ``M'' means that manually labeled segmentation masks by experts are included.}
\label{tab:datasets}
\vspace{-25pt} 
\end{table}



Both datasets (see Table \ref{tab:datasets}) are split into training, validation, and test set. 
To assure a fair analysis of the model's performance, frames from the same video will be part of only one subset. This, however, challenges the balancing of class distributions across the splits (\ie, stratification), as different videos contribute different numbers of sequences/frames to the dataset. To this end, we developed a stratification procedure based on histograms, which tries to preserve a given target distribution of classes in all splits of the dataset. 
Given a target distribution of classes (\eg, corpus distribution), 
we initialize the process by adding a randomly selected video 
to each subset (train, validation, and test). 
Next, we calculate the difference between the target distribution and the current subset's distribution and 
 add the video from the corpus that best compensates the difference by its own intrinsic class distribution. This process is iterated until 
the desired dataset size is reached, leading to a stratified dataset. 


This methodology was applied to create both datasets $D1$ for proxy classification, and $D2$ for segmentation. 
Overall, the dataset contains 685 videos from the corpus and 5974 sequences (5775 homogeneous and 199 heterogeneous), respectively. For $D1$, the distribution of train, validation, and test set is 70\%, 15\%, and 15\%, respectively. For each video sequence in $D1$, the center frame was added to the dataset. Instead, for dataset $D2$, the validation and test subset also used the center frame of each segment and, additionally, these were complemented with more random frames due to a lack of identified and annotated heterogeneous material in our corpus. 


\subsection{Experimental Setup}

\subsubsection{Preprocessing.} 

The visual quality in our corpus is partly low due to the technological limitations of the time and different qualities of video digitization stemming from different decades. 
For this reason, we apply contrast stretching to improve its visual quality. This avoids having particularly dark low-contrast frames in the dataset for which object contours can hardly be recognized. 
Additionally, we apply diverse data augmentation strategies to improve the diversity in the training data and, thereby, the model's generalization ability. Specifically, we apply resizing, random horizontal and vertical flips, random rotation, and perspective transformations. Furthermore, we augment brightness, saturation, contrast, and apply Gaussian noise. 

\subsubsection{Proxy classification (RQ1)} focuses on evaluating the model's ability to distinguish between purely \textit{non-photographic} and purely \textit{photographic} content. To answer RQ1, we use dataset $D1$ containing homogeneous P and NP frames enriched with global labels derived from sequence annotations. 
We use accuracy and F1-score to measure the classification model's performance. 
As described in Sections~\ref{subsec:method_classification} and \ref{subsec:method_dataS}, our primary focus is on finetuning pre-trained models based on Vision Transformer (ViT). 
As baselines, we finetune state-of-the-art models such as ResNet \cite{7780459}, EfficentNet \cite{tan2019efficientnet}, Xception \cite{8099678} and Inception \cite{szegedy2015going}. For all models, we build upon pre-trained weights from the \textit{timm} library \cite{rw2019timm}. 

We optimize different hyperparameters such as the batch size and learning rate. Additionally, we investigate the impact of above mentioned preprocessing methods, \ie, contrast stretching and data augmentation. 
To avoid overfitting, we apply early stopping depending on the validation F1-score. We use the \textit{cross entropy} as loss function and Adam optimizer. A learning rate scheduler is applied as a regularization method, \ie, reduce learning rate on plateau. 
Hyperparameter optimization is conducted via Bayesian optimization provided by the Optuna library \cite{akiba2019optuna}. The optimal ViT model was achieved with a batch size of 6, learning rate of $1 \times 10^{-5}$, and a preprocessing that includes contrast stretching and data augmentation for color frames of 518x518 pixels.

\subsubsection{Boundary cases (RQ1.1)} that are confusing to the classifier and potentially difficult to classify even to human experts, are of particular interest to film scholars to investigate the full spectrum and complexity of hybrid compositions in animation. Thus, we identify uncertain and false predictions by the proxy classifier and investigate them qualitatively. The grounding of the model's decisions is verified via GradCAM~\cite{selvaraju2020grad}, which generates attribution maps at frame level.

\subsubsection{Segmentation (RQ2, RQ2.1, RQ2.2)} aims to investigate the feasibility of segmenting hybrid compositions from proxy segmentation masks and homogeneous P and NP frames. As described in Section~\ref{subsec:method_segmentation}, we finetune the SegFormer segmentation model using the smallest available variant of pre-trained weights with ImageNet-1k (\textit{nvidia/mit-b0}). Additionally, we keep some of the original hyperparameters from~\cite{xie2021segformer}, such as the learning rate of $6 \times 10^{-5}$, and the optimizer AdamW. We adapt the image preprocessing methodology from the original paper by using the existing image processor from HuggingFace, \ie, \textit{SegformerImageProcessor},  
which normalizes, randomly crops, and resizes images and segmentation maps to 512x512 pixels. We employ intersection-over-union (IoU) as performance metric. 
Early stopping is applied during training, monitored by the mean IoU, \ie, the IoU averaged over both classes P and NP. To investigate RQ2.2 (Section~\ref{sec:introduction}), we train the SegFormer with and without heterogeneous frames as well as with and without proxy masks and compare the results.

\section{Results}
\label{sec:results}
\vspace{-5pt} 
\subsection{Quantitative Results}
\subsubsection{Proxy classification task (RQ1).} 
Here, we investigate the classification performance obtained in the proxy task of classifying P vs. NP frames and thereby answer RQ1. To this end, we evaluate two approaches considering the ViT-based architecture from Section~\ref{subsec:method_classification}. In the first approach, we optimized the model by feeding the MLP with the CLS token embedding produced by the ViT backbone (row 1 in \cref{tab:classification_results}). In the second approach, we feed the MLP with the centroid representation $\Bar{H}$ as described in Section~\ref{subsec:method_classification} (row 2 in \cref{tab:classification_results}). 
The employed dataset for these experiments is $D1$, \ie, $D1_{NP+P}^{train}$, $D1_{NP+P}^{val}$, and $D1_{NP+P}^{test}$. 


\vspace{-5pt}
\begin{table}[]
\centering
\begin{tabular}{cccccccc}
\hline
\multicolumn{1}{|c|}{\multirow{2}{*}{\textbf{Network}}} &
  \multicolumn{1}{c|}{\multirow{2}{*}{\textbf{Variant}}} &
  \multicolumn{2}{c|}{\textbf{Training}} &
  \multicolumn{2}{c|}{\textbf{Validation}} &
  \multicolumn{2}{c|}{\textbf{Test}} \\ \cline{3-8} 
\multicolumn{1}{|c|}{} &
  \multicolumn{1}{c|}{} &
  \multicolumn{1}{c|}{\textbf{Acc.}} &
  \multicolumn{1}{c|}{\textbf{F1-score}} &
  \multicolumn{1}{c|}{\textbf{Acc.}} &
  \multicolumn{1}{c|}{\textbf{F1-score}} &
  \multicolumn{1}{c|}{\textbf{Acc.}} &
  \multicolumn{1}{c|}{\textbf{F1-score}} \\ \hline
\multicolumn{1}{|c|}{ViT} &
  \multicolumn{1}{c|}{CLS Token} &
  \multicolumn{1}{c|}{0.9557} &
  \multicolumn{1}{c|}{0.9554} &
  \multicolumn{1}{c|}{0.9436} &
  \multicolumn{1}{c|}{0.9429} &
  \multicolumn{1}{c|}{0.9505} &
  \multicolumn{1}{c|}{0.9505} \\ \hline
\multicolumn{1}{|c|}{ViT} &
  \multicolumn{1}{c|}{Centroid Represent.} &
  \multicolumn{1}{c|}{0.9529} &
  \multicolumn{1}{c|}{0.9526} &
  \multicolumn{1}{c|}{\textbf{0.9540}} &
  \multicolumn{1}{c|}{\textbf{0.9535}} &
  \multicolumn{1}{c|}{\textbf{0.9574}} &
  \multicolumn{1}{c|}{\textbf{0.9574}} \\ \hline
\end{tabular}
\caption{Classification results for the proxy task of classifying photographic vs. non-photographic frames in a binary fashion.}
\label{tab:classification_results}
\vspace{-20pt} 
\end{table}
%
The results show high performance for both ViT-based classifiers in differentiating P vs. NP content at frame level, demonstrating their ability to learn suitable features for the task. The proposed centroid representation slightly outperforms the CLS token approach on both validation and test set.\footnote{Significance not tested due to insufficient number of iterations.} ViT models further notably improve upon the baseline models (\ie, ResNet, EfficentNet, Xception and Inception, not shown in the table for brevity). Among the baseline methods, Inception obtained the highest validation performance of f1=0.9356. This makes ViT the most promising option for solving the proxy task. For ViT, both variants achieve a maximum F1-score close to 95\% in validation and test. 
This also suggests effective 
generalization to unseen data. 



\subsubsection{Image segmentation (RQ2).} In our segmentation experiments, we aim to demonstrate and quantify the effectiveness of including proxy masks generated by our proxy classifier (based on ViT) as supervising signal in the semi-supervised training of our segmentation model (RQ2). For this purpose, we perform two series of segmentation experiments A and B. The employed dataset is $D2$ (see \cref{tab:datasets}), and SegFormer (see Section~\ref{subsec:method_segmentation}) serves as segmentation model.

\textbf{Experiment A: training with only homogeneous frames}:
This experiment aims to establish a baseline for training with only homogeneous frames, \ie, fully NP and fully P frames with their corresponding homogeneous segmentation masks composed of just zeros (NP) or ones (P). As \cref{tab:segmentation_results_A} shows, SegFormer yields outstanding results with homogeneous images, achieving a mean IoU of 89.13\% in validation 
and 83.23\% in the test set, which indicates that our segmentation model has correctly learned the most representative features for homogeneous images, demonstrating good generalization performance.

\vspace{-10pt} 
\begin{table}[]
\centering
\begin{tabular}{ccccccccc}
\multicolumn{9}{c}{\textbf{Experiment A: Baseline on homogeneous frames}} \\ \cline{3-9} 
\multicolumn{2}{c|}{\textbf{}} &
  \multicolumn{1}{c|}{\textbf{Train}} &
  \multicolumn{2}{c|}{\textbf{Validation}} &
  \multicolumn{4}{c|}{\textbf{Test}} \\ \hline 
\multicolumn{2}{|c|}{\textbf{Dataset}} &
  \multicolumn{1}{c|}{$D2_{NP+P}^{train}$} &
  \multicolumn{1}{c|}{$D2_{NP+P}^{val}$} &
  \multicolumn{1}{c|}{$D2_{NP+P+M}^{val}$} &
  \multicolumn{1}{c|}{$D2_{NP+P}^{test}$} &
  \multicolumn{1}{c|}{$D2_{NP+P+M}^{test}$} &
  \multicolumn{1}{|c|}{} &
  \multicolumn{1}{|c|}{$D2_{M}^{test}$} \\ \cline{1-7} \cline{9-9}
\multicolumn{2}{|c|}{\textbf{Mean IoU}} &
  \multicolumn{1}{c|}{0.9878} &
  \multicolumn{1}{c|}{0.8913} &
  \multicolumn{1}{c|}{0.6202} &
  \multicolumn{1}{c|}{0.8323} &
  \multicolumn{1}{c|}{0.5989} &
  \multicolumn{1}{|c|}{} &
  \multicolumn{1}{|c|}{0.2462}\\ \cline{1-7} \cline{9-9}
\multicolumn{1}{|c|}{\multirow{2}{*}{\textbf{IoU per class}}} &
  \multicolumn{1}{c|}{\textbf{NP}} &
  \multicolumn{1}{c|}{0.9877} &
  \multicolumn{1}{c|}{0.8877} &
  \multicolumn{1}{c|}{0.5918} &
  \multicolumn{1}{c|}{0.8249} &
  \multicolumn{1}{c|}{0.5546}  &
  \multicolumn{1}{|c|}{} &
  \multicolumn{1}{|c|}{0.0921}\\ \cline{2-7} \cline{9-9} 
\multicolumn{1}{|c|}{} &
  \multicolumn{1}{c|}{\textbf{P}} &
  \multicolumn{1}{c|}{0.9879} &
  \multicolumn{1}{c|}{0.8949} &
  \multicolumn{1}{c|}{0.6487} &
  \multicolumn{1}{c|}{0.8398} &
  \multicolumn{1}{c|}{0.6432}  &
  \multicolumn{1}{|c|}{} &
  \multicolumn{1}{|c|}{0.4003}\\ \hline
\end{tabular}%
\caption{Baseline results for training the segmentation model from only homogeneous frames, \ie, without proxy masks.}
\label{tab:segmentation_results_A}
\vspace{-30pt} 
\end{table}


The main question that arises with regard to these baseline results is, how well this trained segmentation model can cope with heterogeneous frames that it never saw during training. To this end, we also evaluate the model with 
the validation and test subsets composed of homogeneous and heterogeneous data ($D2_{NP+P+M}$). As expected, the performance decreases compared to that obtained with only homogeneous data. Specifically, metrics degrade, with a mean IoU of 59.89\% for the test set. 
These results show that training on only homogeneous frames is insufficient to obtain a segmentation of hybrid frames. We use these results as a reference to measure the impact of including proxy masks in the training process (Experiment B, RQ2).

\textbf{Experiment B: Training with homogeneous and heterogeneous fra-\\mes}: Here, 
we differentiate 
between three sub-experiments, B.0, B.1, and B.2, which differ in the type of supervisory information used in training. 



\begin{table}[ht]
\begin{subtable}[t]{\textwidth}
\centering
\resizebox{\columnwidth}{!}{%
\begin{tabular}{cccccccccc}
\multicolumn{10}{c}{\textbf{Experiment B.0: Supervised baseline}} \\ \cline{3-10} 
\multicolumn{2}{c|}{} &
  \multicolumn{2}{c|}{\textbf{Train}} &
  \multicolumn{2}{c|}{\textbf{Validation}} &
  \multicolumn{4}{c|}{\textbf{Test}} \\ \hline
\multicolumn{2}{|c|}{\textbf{Dataset}} &
  \multicolumn{1}{c|}{$D2_{NP+P}^{train}$} &
  \multicolumn{1}{c|}{$D2_{NP+P+M}^{train}$} &
  \multicolumn{1}{c|}{$D2_{NP+P}^{val}$} &
  \multicolumn{1}{c|}{$D2_{NP+P+M}^{val}$} &
  \multicolumn{1}{c|}{$D2_{NP+P}^{test}$} &
  \multicolumn{1}{c|}{$D2_{NP+P+M}^{test}$} &
  \multicolumn{1}{c|}{} &
  \multicolumn{1}{c|}{$D2_{M}^{test}$}
  \\ \cline{1-8} \cline{10-10}
\multicolumn{2}{|c|}{\textbf{Mean IoU}} &
  \multicolumn{1}{c|}{0.9981} &
  \multicolumn{1}{c|}{0.9504} &
  \multicolumn{1}{c|}{0.8453} &
  \multicolumn{1}{c|}{0.7965} &
  \multicolumn{1}{c|}{0.8245} &
  \multicolumn{1}{c|}{\textbf{0.7687}} &
  \multicolumn{1}{c|}{} &
  \multicolumn{1}{c|}{0.5361} \\ \cline{1-8} \cline{10-10}
\multicolumn{1}{|c|}{\multirow{2}{*}{\textbf{IoU per category}}} &
  \multicolumn{1}{c|}{\textbf{NP}} &
  \multicolumn{1}{c|}{0.9981} &
  \multicolumn{1}{c|}{0.9406} &
  \multicolumn{1}{c|}{0.8402} &
  \multicolumn{1}{c|}{0.7543} &
  \multicolumn{1}{c|}{0.8171} &
  \multicolumn{1}{c|}{0.7153}  &
  \multicolumn{1}{c|}{} &
  \multicolumn{1}{c|}{0.2621} \\ \cline{2-8} \cline{10-10}
\multicolumn{1}{|c|}{} &
  \multicolumn{1}{c|}{\textbf{P}} &
  \multicolumn{1}{c|}{0.9981} &
  \multicolumn{1}{c|}{0.9602} &
  \multicolumn{1}{c|}{0.8504} &
  \multicolumn{1}{c|}{0.8387} &
  \multicolumn{1}{c|}{0.8318} &
  \multicolumn{1}{c|}{0.8222}  &
  \multicolumn{1}{c|}{} &
  \multicolumn{1}{c|}{0.8101} \\ \hline
\end{tabular}%
}
\caption{\textbf{Experiment B.0} is a completely supervised experiment were training is conducted with full pixel-accurate ground truth masks manually generated for training and validation sets ($D2_{NP+P+M}^{train}$ and $D2_{NP+P+M}^{val}$). We use this ideal setting to show what is feasible in the best case.}
\label{tab:segmentation_results_B0}
\end{subtable}

\begin{subtable}[t]{\textwidth}
\centering
\resizebox{\columnwidth}{!}{%
\begin{tabular}{cccccccccc}
\multicolumn{10}{c}{\textbf{Experiment B.1: Semi-supervised training guided by proxy masks}} \\ \cline{3-10} 
\multicolumn{2}{c|}{} &
  \multicolumn{2}{c|}{\textbf{Train}} &
  \multicolumn{2}{c|}{\textbf{Validation}} &
  \multicolumn{4}{c|}{\textbf{Test}} \\ \hline
\multicolumn{2}{|c|}{\textbf{Dataset}} &
  \multicolumn{1}{c|}{$D2_{NP+P}^{train}$} &
  \multicolumn{1}{c|}{$D2_{NP+P+Proxy}^{train}$} &
  \multicolumn{1}{c|}{$D2_{NP+P}^{val}$} &
  \multicolumn{1}{c|}{$D2_{NP+P+M}^{val}$} &
  \multicolumn{1}{c|}{$D2_{NP+P}^{test}$} &
  \multicolumn{1}{c|}{$D2_{NP+P+M}^{test}$} & 
  \multicolumn{1}{c|}{} &
  \multicolumn{1}{c|}{$D2_{M}^{test}$} \\ \cline{1-8} \cline{10-10}
\multicolumn{2}{|c|}{\textbf{Mean IoU}} &
  \multicolumn{1}{c|}{0.9997} &
  \multicolumn{1}{c|}{0.8890} &
  \multicolumn{1}{c|}{0.7801} &
  \multicolumn{1}{c|}{0.7059} &
  \multicolumn{1}{c|}{0.7810} &
  \multicolumn{1}{c|}{\textbf{0.7470}} &
  \multicolumn{1}{c|}{} &
  \multicolumn{1}{c|}{0.5082} \\ \cline{1-8} \cline{10-10}
\multicolumn{1}{|c|}{\multirow{2}{*}{\textbf{IoU per category}}} &
  \multicolumn{1}{c|}{\textbf{NP}} &
  \multicolumn{1}{c|}{0.9797} &
  \multicolumn{1}{c|}{0.8786} &
  \multicolumn{1}{c|}{0.7563} &
  \multicolumn{1}{c|}{0.6291} &
  \multicolumn{1}{c|}{0.7574} &
  \multicolumn{1}{c|}{0.6704} &
  \multicolumn{1}{c|}{} &
  \multicolumn{1}{c|}{0.1682} \\ \cline{3-8} \cline{10-10}
\multicolumn{1}{|c|}{} &
  \multicolumn{1}{c|}{\textbf{P}} &
  \multicolumn{1}{c|}{0.9801} &
  \multicolumn{1}{c|}{0.8994} &
  \multicolumn{1}{c|}{0.8039} &
  \multicolumn{1}{c|}{0.7826} &
  \multicolumn{1}{c|}{0.8046} &
  \multicolumn{1}{c|}{0.8234} &
  \multicolumn{1}{c|}{} &
  \multicolumn{1}{c|}{0.8482}\\ \hline
\end{tabular}%
}
\caption{\textbf{Experiment B.1} uses proxy masks during training ($D2_{NP+P+Proxy}^{train}$), while for the heterogeneous samples in the validation subset manual ground truth masks are provided ($D2_{NP+P+M}^{val}$).
}
\label{tab:segmentation_results_B1}
\end{subtable}

\begin{subtable}[t]{\textwidth}
\centering
\resizebox{\columnwidth}{!}{%
\begin{tabular}{ccccccccccc}
\multicolumn{11}{c}{\textbf{Experiment B.2: Semi-supervised training and validation guided by proxy masks}} \\ \cline{3-11} 
\multicolumn{2}{c|}{} &
  \multicolumn{2}{c|}{\textbf{Train}} &
  \multicolumn{3}{c|}{\textbf{Validation}} &
  \multicolumn{4}{c|}{\textbf{Test}} \\ \hline
\multicolumn{2}{|c|}{\textbf{Dataset}} &
  \multicolumn{1}{c|}{$D2_{NP+P}^{train}$} &
  \multicolumn{1}{c|}{$D2_{NP+P+Proxy}^{train}$} &
  \multicolumn{1}{c|}{$D2_{NP+P}^{val}$} &
  \multicolumn{1}{c|}{$D2_{NP+P+Proxy}^{val}$} &
  \multicolumn{1}{c|}{$D2_{NP+P+M}^{val}$} &
  \multicolumn{1}{c|}{$D2_{NP+P}^{test}$} &
  \multicolumn{1}{c|}{$D2_{NP+P+M}^{test}$} &
  \multicolumn{1}{c|}{} &
  \multicolumn{1}{c|}{$D2_{M}^{test}$} \\ \cline{1-9} \cline{11-11} 
\multicolumn{2}{|c|}{\textbf{Mean IoU}} &
  \multicolumn{1}{c|}{0.9999} &
  \multicolumn{1}{c|}{0.9275} &
  \multicolumn{1}{c|}{0.8887} &
  \multicolumn{1}{c|}{0.7898} &
  \multicolumn{1}{c|}{0.7261} &
  \multicolumn{1}{c|}{0.8437} &
  \multicolumn{1}{c|}{\textbf{0.7498}} &
  \multicolumn{1}{c|}{} &
  \multicolumn{1}{c|}{0.4722} \\ \cline{1-9} \cline{11-11} 
\multicolumn{1}{|c|}{\multirow{2}{*}{\textbf{IoU per category}}} &
  \multicolumn{1}{c|}{\textbf{NP}} &
  \multicolumn{1}{c|}{0.9999} &
  \multicolumn{1}{c|}{0.9226} &
  \multicolumn{1}{c|}{0.8830} &
  \multicolumn{1}{c|}{0.7660} &
  \multicolumn{1}{c|}{0.6798} &
  \multicolumn{1}{c|}{0.8330} &
  \multicolumn{1}{c|}{0.6917} &
  \multicolumn{1}{c|}{} &
  \multicolumn{1}{c|}{0.1949} \\ \cline{2-9} \cline{11-11} 
\multicolumn{1}{|c|}{} &
  \multicolumn{1}{c|}{\textbf{P}} &
  \multicolumn{1}{c|}{0.9999} &
  \multicolumn{1}{c|}{0.9325} &
  \multicolumn{1}{c|}{0.8944} &
  \multicolumn{1}{c|}{0.8136} &
  \multicolumn{1}{c|}{0.7723} &
  \multicolumn{1}{c|}{0.8545} &
  \multicolumn{1}{c|}{0.8079} &
  \multicolumn{1}{c|}{} &
  \multicolumn{1}{c|}{0.7495} \\ \hline
\end{tabular}%
}
\caption{\textbf{Experiment B.2} exclusively trains using proxy masks generated by our classification model in the training and validation sets ($D2_{NP+P+Proxy}^{train}$ and $D2_{NP+P+Proxy}^{val}$).
}
\label{tab:segmentation_results_B2}
\end{subtable}
\caption{Results from Experiment B on homogeneous and heterogeneous frames.}
\label{tab:segmentation_results_B}
\vspace{-25pt} 
\end{table}

Experiment B.0 represents a supervised baseline of the segmentation performance as it pursues a fully supervised learning process using ground truth in the form of manually labeled masks. This supervised baseline is 0.7687 (mean IoU) for the test set $D2_{NP+P+M}^{test}$ containing homogeneous and heterogeneous frames.
In contrast, experiment B.1 incorporates proxy masks in the training while manual masks are used for validation and testing; ultimately, B.2 involves proxy masks in training and validation, testing just on the ground truth. As detailed in \cref{tab:segmentation_results_B}, the semi-supervised segmentation models from B.1 and B.2 have a test performance on the homogeneous material ($D2_{NP+P}^{test}$) that is comparable to that obtained in Experiment A, demonstrating that the inclusion of proxy masks preserves the good performance on fully homogeneous data. Notably, B.2 even outperforms experiment A with a mean IoU of 84\% on the test set (\cref{tab:segmentation_results_A}). 

To answer RQ2.2, semi-supervised training in experiment B improves the segmentation performance of heterogeneous and homogeneous test data, $D2_{NP+P+M}^{test}$, achieving approx. 75\% of mean IoU compared to 60\% from experiment A, which demonstrates that proxy masks strongly improve segmentation performance. The performance of B.1 and B.2 is even close to the fully supervised baseline (B.0) with almost 77\%.



\textbf{Evaluating segmentation performance on purely heterogeneous material (RQ2.2)}: Finally, we investigate the segmentation model's performance exclusively on heterogeneous material, which represents the hybrid compositions we are particularly interested in. To this end, we use the manually labeled test data, $D2_{M}^{test}$. Results in the last column of experiments tables (A: \cref{tab:segmentation_results_A} and B: \cref{tab:segmentation_results_B}), labeled $D2_{M}^{test}$, show that segmentation of hybrid compositions from training with only homogeneous frames (experiment A) does not suffice (mean IoU of 24.62\%). The inclusion of proxy masks notably improves the segmentation performance from Exp. A to a mean IoU of 47.22\% in Exp. B.2, and even to 50.82\% in Exp. B.1. This performance is close to the supervised baseline of 53.61\% that we achieve with fully labeled data in Exp. B.0.
\vspace{-5pt} 
\subsection{Qualitative Results}
In this section, we qualitatively analyze the results from the proxy task and from segmentation and investigate RQ1.1 and RQ2.1.
\vspace{-5pt} 
\subsubsection{Boundary cases in classification (RQ1.1).}\label{subsubsec:bound_classif} 
\begin{figure}[t!]
    \centering
    \begin{subfigure}[b]{0.255\textwidth}
    \includegraphics[width=\textwidth]{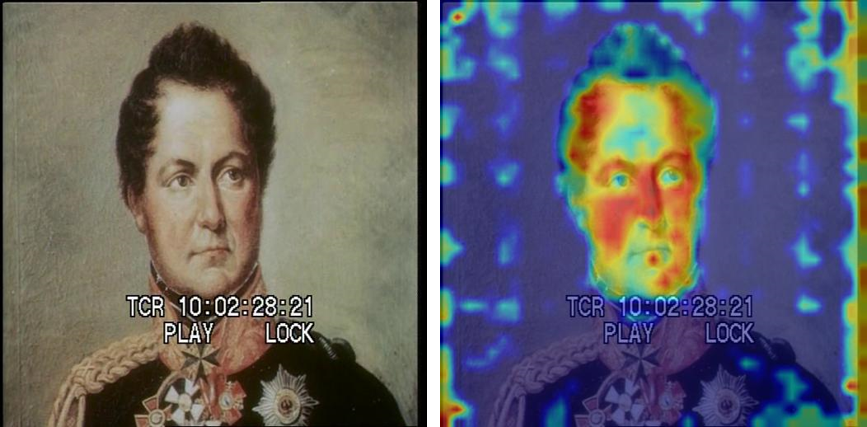}
        \caption{NP; Pred. P(0.57)}
        \label{fig:hist_1}
    \end{subfigure}
    \hspace{0.01cm} 
    \begin{subfigure}[b]{0.255\textwidth}
    \includegraphics[width=\textwidth]{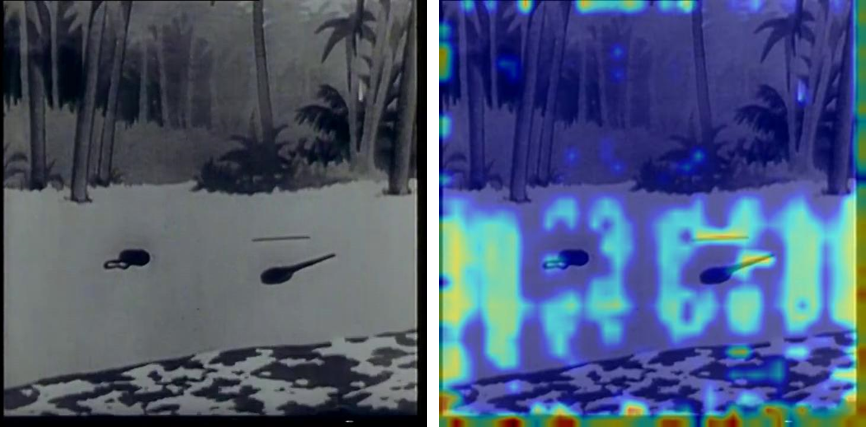}
        \caption{NP; Pred.: P(0.51)}
        \label{fig:hist_2}
    \end{subfigure}
    \hspace{0.01cm} 
    \begin{subfigure}[b]{0.26\textwidth}
        \includegraphics[width=\textwidth]{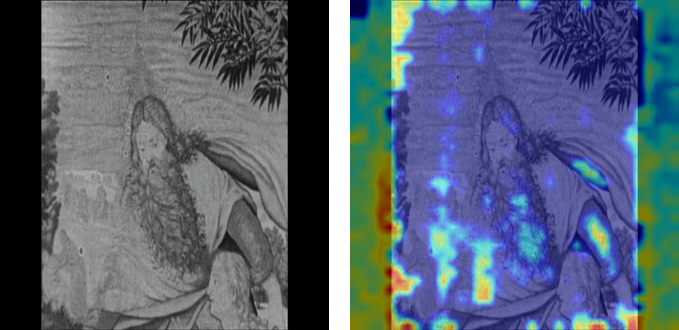}
        \caption{NP; Pred.: P(0.57)}
        \label{fig:hist_3}
    \end{subfigure}

    \vspace{0.1cm}

    \begin{subfigure}[b]{0.255\textwidth}
        \includegraphics[width=\textwidth]{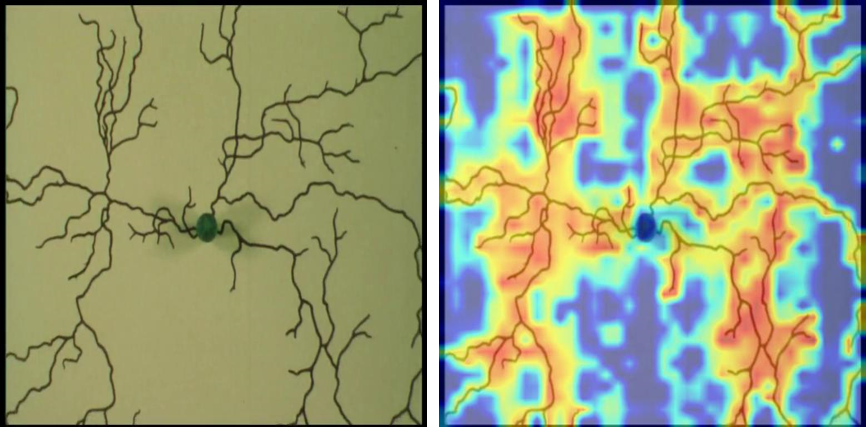}
        \caption{P; Pred.: NP(0.98)}
        \label{fig:hist_4}
    \end{subfigure}
    \hspace{0.01cm}  
    \begin{subfigure}[b]{0.26\textwidth}
        \includegraphics[width=\textwidth]{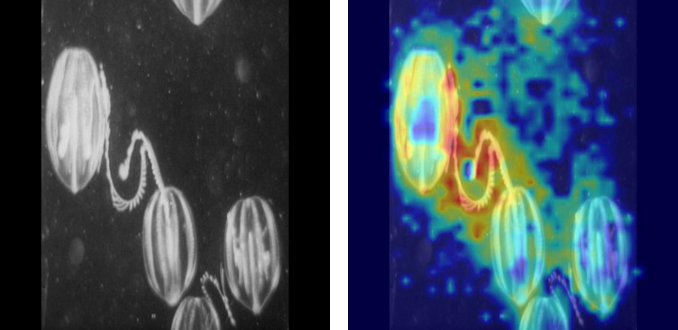}
        \caption{P; Pred.: NP(0.54)}
        \label{fig:hist_5}
    \end{subfigure}
    
    \caption{Boundary cases in the proxy classification task and GradCAM explanations (red high, blue low attribution for predicted class). For each image, we provide the true label (P or NP) and the model's prediction with its likelihood in brackets.\\ \textit{Sources}: (a)\cite{Neithardt}, (b)\cite{Hai}, (c)\cite{Polen}, (d)\cite{tibav:10606}, (e)\cite{Quallen}}
    \label{fig:subfigures}
    \vspace{-10pt} 
\end{figure}

To analyze boundary cases and investigate RQ1.1, we apply the GradCAM explainability algorithm \cite{selvaraju2020grad} on misclassified samples from $D1$ (test and validation set). GradCAM generates attribution masks, \ie, heatmaps, which highlight the most influential pixels for the model's predictions. 
We found boundary cases in the form of very realistic drawings, specifically portraits, where the model assumes photographic content in the face area (\cref{fig:hist_1}). Another problematic cases are drawings that tend to look like old, noisy black and white photographic images, making them challenging for even humans to recognize as NP frame (\cref{fig:hist_2}). Other difficult cases include photographic images of real objects artistically arranged, resulting in abstract compositions of strokes, lines or shadows about which the model is unsure or even totally confused (\cref{fig:hist_4}). Similarly, abstract scientific photographic content is often considered NP, as shown in \cref{fig:hist_5}. Lastly, monochrome backgrounds and black borders confuse the network as they appear in both classes (\eg, \cref{fig:hist_3}). 

\vspace{-15pt} 
\subsubsection{Quality of proxy masks (RQ2.1).} 
\begin{figure}[ht]
    \centering
    \begin{minipage}{0.48\textwidth}
        \centering
        \begin{subfigure}{\textwidth}
            \centering
            \includegraphics[width=\linewidth]{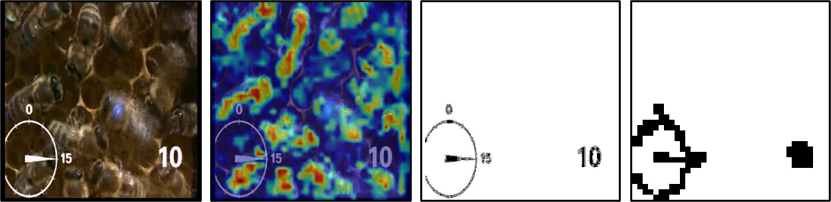}
            \caption{\textit{Source}: \cite{tibav:23086}}
            \label{fig:seg_1}
        \end{subfigure}
        
        \begin{subfigure}{\textwidth}
            \centering
            \includegraphics[width=\linewidth]{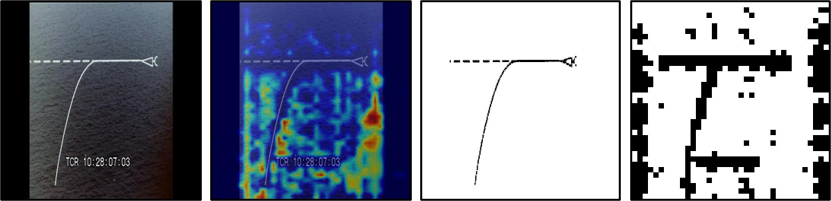}
            \caption{\textit{Source}: \cite{Torpedo}}
            \label{fig:seg_2}
        \end{subfigure}
        

    \end{minipage}
    \hfill
    \begin{minipage}{0.48\textwidth}
        \centering
        \begin{subfigure}{\textwidth}
            \centering
            \includegraphics[width=\linewidth]{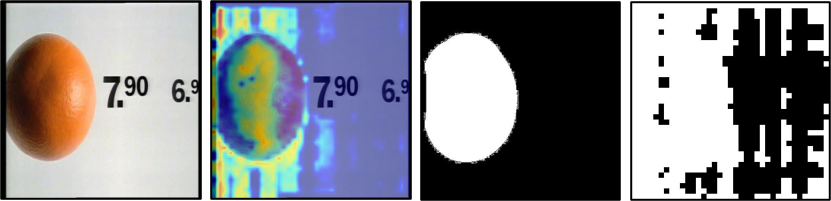}
            \caption{\textit{Source}: \cite{Werbung}}
            \label{fig:seg_3}
        \end{subfigure}

        
        \begin{subfigure}{\textwidth}
            \centering
            \includegraphics[width=\linewidth]{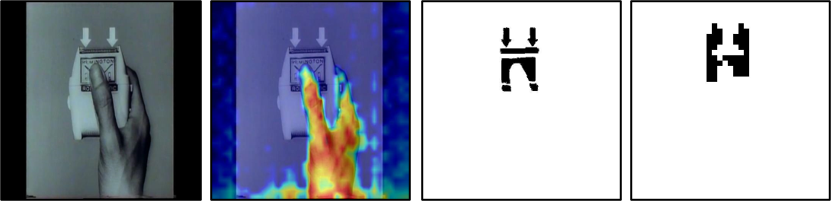}
            \caption{\textit{Source}: \cite{Remington}} 
            \label{fig:seg_7}
        \end{subfigure}
        
    \end{minipage}
    \caption{Hybrid frames together with GradCAM explanations (for the class P), the ground truth (black NP, white P), and proxy segmentation masks. 
    }
    \label{fig:seg_imgs}
    \vspace{-10pt} 
\end{figure}

\cref{fig:seg_imgs} shows examples of hybrid frames together with GradCAM explanations, ground truth and proxy segmentation masks. Proxy masks have surprisingly good results even in difficult cases, see, \eg, \cref{fig:seg_1} (fine contours) and
~\ref{fig:seg_7} (two small arrows and hard to recognize illustrated content on the device). As previously observed black borders confuse the model (\cref{fig:seg_2}), which is however not always the case (\cref{fig:seg_7}). Monochrome  (\cref{fig:seg_3}) and low-contrast frames (\cref{fig:seg_2}) may confuse the model.

The qualitative results highlight another important finding, namely that the segmentation task is strongly unbalanced concerning the area covered by the P and NP classes in the hybrid material, with NP being the minority class. 
 This imbalance challenges the segmentation model overall, which explains the 
imbalance in IoU per class for heterogeneous material ($D2_{M}^{test}$) in Exp. A (\cref{tab:segmentation_results_A}) and Exp. B (\cref{tab:segmentation_results_B}).
Finally, RQ2.1 can be answered positively.

\vspace{-10pt}
\subsubsection{Segmentation results.}

\begin{figure}[ht]
    \centering
    
    \begin{subfigure}[b]{0.282\textwidth}
        \centering\includegraphics[width=\textwidth]{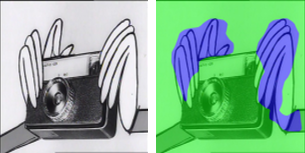}
        \caption{\textit{Source}: \cite{Werbung}}
        \label{fig:Seg_sub1}
    \end{subfigure}
    \hspace{0.005\textwidth}
    \begin{subfigure}[b]{0.32\textwidth}
        \centering\includegraphics[width=\textwidth]{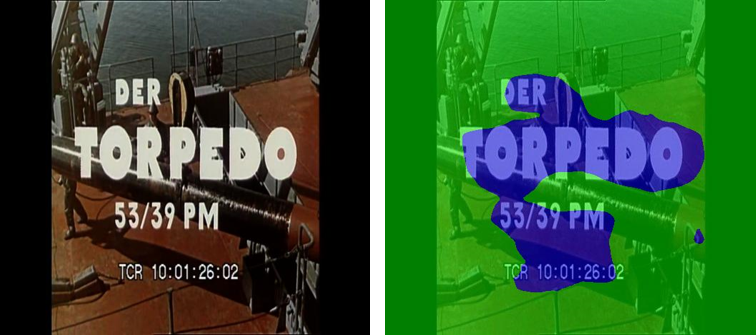}
        \caption{\textit{Source}: \cite{Torpedo}}
        \label{fig:Seg_sub2}
    \end{subfigure}
    \hspace{0.005\textwidth}
    \begin{subfigure}[b]{0.28\textwidth}
        \centering       \includegraphics[width=\textwidth]{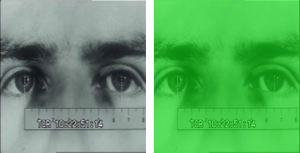}
        \caption{\textit{Source}: \cite{Grundlage}}
        \label{fig:Seg_sub3}
    \end{subfigure}
    
    \caption{Segmentation results on hybrid frames (P: green; NP: blue).
    }
    \label{fig:figura_con_subfiguras}
    \vspace{-25pt} 
\end{figure}

\cref{fig:figura_con_subfiguras} presents segmentation results for hybrid frames. In the first example (\cref{fig:Seg_sub1}), the fingers holding the camera are correctly labeled as NP while the arms are not segmented. In \cref{fig:Seg_sub2} the overlaid text is correctly labeled as NP and the background is correctly detected as P. In \cref{fig:Seg_sub3} the ruler represents an undiscovered NP element that is particularly hard to detect even for humans and shows the limitations of automated analysis.

\section{Conclusion}
\label{sec:conclusion}

This work contributes a first quantitative computer vision-supported analysis of hybrid visual compositions in animation, with a special focus on ephemeral film. We propose a weakly and semi-supervised approach for segmenting hybrid frames into their building blocks, namely photographic and non-photographic content. We demonstrate that with only a minimum of ground truth information (just sequence level annotations, no segmentation masks) we can achieve  segmentation performance close to a completely supervised baseline. Our approach enables to automatically identify hybrid compositions in unannotated corpora, identify boundary cases and thereby supports the stylistic analysis of 
film scholars. Future topics are the compilation of a large-scale dataset of hybrid material and the development of strategies to cope with class spatial imbalance. 


\section{Acknowledgements}
We would like to thank project members Kristina Schmiedl, Mahboobeh Mohammadzaki, and Clemens Baumann for their contributions. AniVision is a collaboration between the St Pölten University of Applied Sciences and the University of Tübingen, funded by the Austrian Science Fund (FWF) (project no.: I5592-G) and the Deutsche Forschungsgemeinschaft (DFG) (project no.: 468856086).

%
%
\bibliographystyle{splncs04}
\bibliography{main}

\end{document}